\documentclass{article}
    \PassOptionsToPackage{numbers, compress}{natbib}
    \usepackage[preprint]{neurips_2019}

\usepackage[utf8]{inputenc} % allow utf-8 input
\usepackage[T1]{fontenc}    % use 8-bit T1 fonts
\usepackage{hyperref}       % hyperlinks
\usepackage{url}            % simple URL typesetting
\usepackage{graphicx}
\usepackage{booktabs}       % professional-quality tables
\usepackage{amsfonts}       % blackboard math symbols
\usepackage{nicefrac}       % compact symbols for 1/2, etc.
\usepackage{microtype}      % microtypography
\title{Distributed low-precision Training Without Mixed Precision}
\author{%
Zehua Cheng$^{1, 2, *}$, Weiyang Wang$^{1}$\thanks{Equal Contributions}, Yan Pan$^1$, Thomas Lukasiewicz$^2$\\
$^1$SnowCloud.ai\\
$^2$Department of Computer Science, University of Oxford\\
\texttt{\{limber | weiyang.wang\}@snowcloud.ai}
}

\begin{document}
\maketitle

\begin{abstract}
  Low-precision training is one of the most popular strategies for deploying deep neural models on hardware with resource constraints.
  Fixed-point implementation of deep models has the potential to alleviate complexities and facilitate potential deployment on embedded hardware. However, most low-precision training solutions are based on a mixed precision strategy. In this paper, we present an ablation study on different low-precision training strategies and propose a solution for IEEE FP-16 format throughout the training process. We demonstrate the effectiveness of our solution with ResNet50 on the ImageNet-full dataset using a cluster with 128 GPUs. 
  The experimental results show that it is not essential to use FP32 format to train deep models. 
%   We have viewed that communication cost reduction, model compression, and large-scale distributed training are three coupled problems.
 \end{abstract}
 \section{Introduction}
Deep convolutional networks have widely been applied to the non-trivial machine learning problems in computer vision task~\cite{lecun2015deep,krizhevsky2012imagenet}, and there has been a gradual advance in the model representation. 
 However, complex models are computationally intensive, which makes it hard to deploy on embedded systems.
 To address this issue, there has been a surge of interests recently in reducing the model complexity of the deep model, including light-weight deep learning models utilizing computation/memory efficient operations, exemplified by MobileNet~\cite{howard2017mobilenets}, ShuffleNet~\cite{zhang2018shufflenet}, quantization~\cite{courbariaux2015binaryconnect,zhu2016trained,jacob2018quantization,rastegari2016xnor} and pruning~\cite{han2015learning,he2017channel}. Among these approaches, quantization that reduce the precision requirements for the weights and activations by representing network weights with low-precision (lower then FP32), thus yielding highly compact deep models compared to their floating-point counterparts.
 
 Meanwhile, the quantization not only requires less working memory and cache but accelerate the computation speed and cuts down the energy consumption of the computational device. \cite{de2018high} has proved the feasibility of low-precision training in a well-conditioned situation. It has proved that reduce the precision of the data would not significantly change the distribution of the data, so it is feasible to replace float32 data with fixed-point data, which has dramatically reduced performance and space consumption.
 The details of the energy consumption of the different operations in different precision are presented in Table~\ref{table:energy_consumption}. However, it is necessary to adjust the order values to avoid overflow dynamically. 
 Unfortunately, the statistical effects of low-precision computation during training are still not well understood. Therefore, if the pure low-precision computation is used throughout training, it is often challenging to match the statistical accuracies of traditional, higher-precision hardware architectures. Hence, the mixed-precision strategy is becoming popular. \cite{courbariaux2015binaryconnect} proposed training with binary weights, all other tensors and arithmetic were in full precision. \cite{courbariaux2016binarized} extended that work also to binarize the activations, but gradients were stored and computed in single precision. \cite{hubara2017quantized} considered quantization of weights and activations to $2, 4$ and $6$ bits, gradients were real numbers. Besides the mixed-precision training strategy, most of the works are not adopted on large scale datasets like ImageNet~\cite{deng2009imagenet} and very deep models like ResNet\cite{He2016}. Fixed precision training is a strategy that aims to tackle the challenge of the above problems. Currently, the state-of-the-art mixed-precision training strategy like~\cite{micikevicius2017mixed} using the strategy of maintaining a master copy of weights in FP32, loss-scaling that minimizes gradient values becoming zeros, and FP16 arithmetic with accumulation in FP32 indicating that strategy is not a full FP16 training process. This strategy has significantly increased the training cost and hard to implement on a large-scale distributed system. 
 What is more, a fully FP16 training strategy could also take advantage of the NVIDIA Pascal GPU, which support to accelerate the whole training process twice~\cite{ho2017exploiting}. With mentioned above disadvantages, we viewed that the mixed-precision training strategy is an essential exploration to achieve pure FP16 training.
 % To this extent, we viewed that full FP16 training could also consider as an accelerating deep model training strategy.
 
 \begin{table}[!ht]
  % 整型数据在资源消耗上大大低于浮点数
  \caption{The energy consumption of different operations in different precision\label{table:energy_consumption}}
  \centering
  \begin{tabular}{c|c|c|c}
  Precision&Operation & Energy(pJ) & Area($\mu m^2$) \\\hline
  FP16&addition&0.4&1,360\\
  FP32&addition&0.9&4,184\\
  FP16&multiplication&1.1&1,640\\
  FP32&multiplication&3.7&7,700\\
  \end{tabular}
 \end{table}
 
 Training on FP16 reduces memory consumption by half compared with FP32. The main challenge of non-full FP16 training is that the FP16 has a narrower dynamic range than FP32. Some operations like Batch Normalization~\cite{ioffe2015batch} require the computation of the sum of squares, square-root, and reciprocal operations to avoid zero variance, which requires high precision and a broad dynamic range. Thus, some works avoid to use Batch Normalization layer~\cite{wu2018training} or kept the precision of the parameters unchanged~\cite{zhou2016dorefa} to avoid facing this challenge. To tackle this problem, \cite{banner2018scalable} has proposed Range Batch Normalization that normalizes inputs by the range of the input distribution. However, all the existing quantization solutions are based on 1-bit inference, which is insensitive to the gamma in the Batch Normalization layer before each ReLU. 
 
 In this work, we proposed to easily swap the position of the Gamma/Beta to solve this problem, which is a more straightforward approach to solve this problem. We also developed a low-precision training as an approach to accelerating the deep neural networks (DNN) training approach. 
 There is no doubt that low-precision training would cut down the budget of the GPU storages, so scaling full low-precision training is also appealing. However, even for the traditional FP32 training model, scaling ImageNet-1k training would still waste great computational resources, communication bandwidth. Moreover, end-to-end training would be more efficient to train the models. \cite{you2018imagenet} proposed Layer-wise Adaptive Rate Scaling(LARS) to address this issue.
 
 Our specific contributions are:
 \begin{enumerate}
  \item Proposed a fully FP16 training without FP32 throughout the training process on ImageNet while maintaining model accuracy. To our knowledge, our proposed strategy is the first fully FP16 training solution without sacrificing accuracy.
  \item We proved the scalability of our proposed FP16 training strategy over the large-scale distributed deep learning training cluster without accuracy sacrificed.
 \end{enumerate}
 \section{Related Works}
 Reduced precision methods for deep learning training is an approach to improving compute efficiency by reducing the precision requirements for the weights and activations by representing network weights with low-precision. Quantized weights significantly reduce memory size and access bandwidth, improve power efficiency exploit hardware-friendly bitwise operations, and accelerate inference throughput.
 Moreover, since the bandwidth is the major bottleneck of the distributed system~\cite{you2018imagenet}, the mitigated bandwidth would also help to improve the utilization of the computational resources, thus accelerate the model training process. \cite{gupta2015deep} has demonstrated that deep neural networks can be trained with minimal loss in accuracy, using 16-bit fixed-point representation. However, most of the low-precision training strategies were not using deep models like ResNet~\cite{He2016} and tested on large-scale datasets like ImageNet~\cite{deng2009imagenet} because most low-precision training strategies are highly unstable. 
 Accumulation of errors will collapse the accuracy of the whole model with the updating process by back propagation.
 It is challenging to deal with low-precision weights and stimulate researchers' interest in new training methods. The main challenge lies when the learning rate is very small, the stochastic gradient method updates the weight parameters. \cite{banner2018scalable} applied their proposed algorithm to deep convolutional models where they analyzed the quantization sensitivity of the network for each layer and then manually decide the quantization bitwidths. \cite{vanhoucke2011improving} quantized the weights and activations of pre-trained deep networks using 8-bit fixed-point representation to improve inference speed.
 After each training iteration, the weights are binarized/discretized and rounded up, which leads to training stagnation~\cite{courbariaux2016binarized}. Some updating gradients are not helpful for the training process~\cite{cheng2019bandwidth}. Thus the straightforward approach that quantizing weights with a rounding procedure yields unsatisfactory results when weights are represented in low-precision numbers. 
 The other approaches address this problem by using mixed-precision training strategies that using full precision and single precision together during the training procedure~\cite{mishra2017wrpn}.
 ~\cite{micikevicius2017mixed} that maintaining a master copy of weights in FP32, loss-scaling that minimizes gradient values becoming zeros, and FP16 arithmetic with accumulation in FP32. 
 
 % \subsection{Accelerating Large-Scale Distributed Training}
 % HKM 1H\cite{goyal2017accurate}
 % The main challenges of accelerating deep learning training procedure is bandwidth communication. The goal of designing an acceleration solution for the large scale distributed training system is to increase the computation to communication ratio.
 % The state-of-the-art results of scaling ImageNet-1k training have used the synchronous stochastic gradient descent method~\cite{goyal2017accurate}. Layer-wise Adaptive Rate Scaling (LARS) algorithm~\cite{you2018imagenet} is a strategy to dramatically scale the batch size and use the massive computing resources efficiently. The motivation of the design of the LARS is to cut down the communication ratio.
 
 \section{Precision Standard}
 
 \begin{figure}[!ht]
  \centering
  \includegraphics[width=0.85\textwidth]{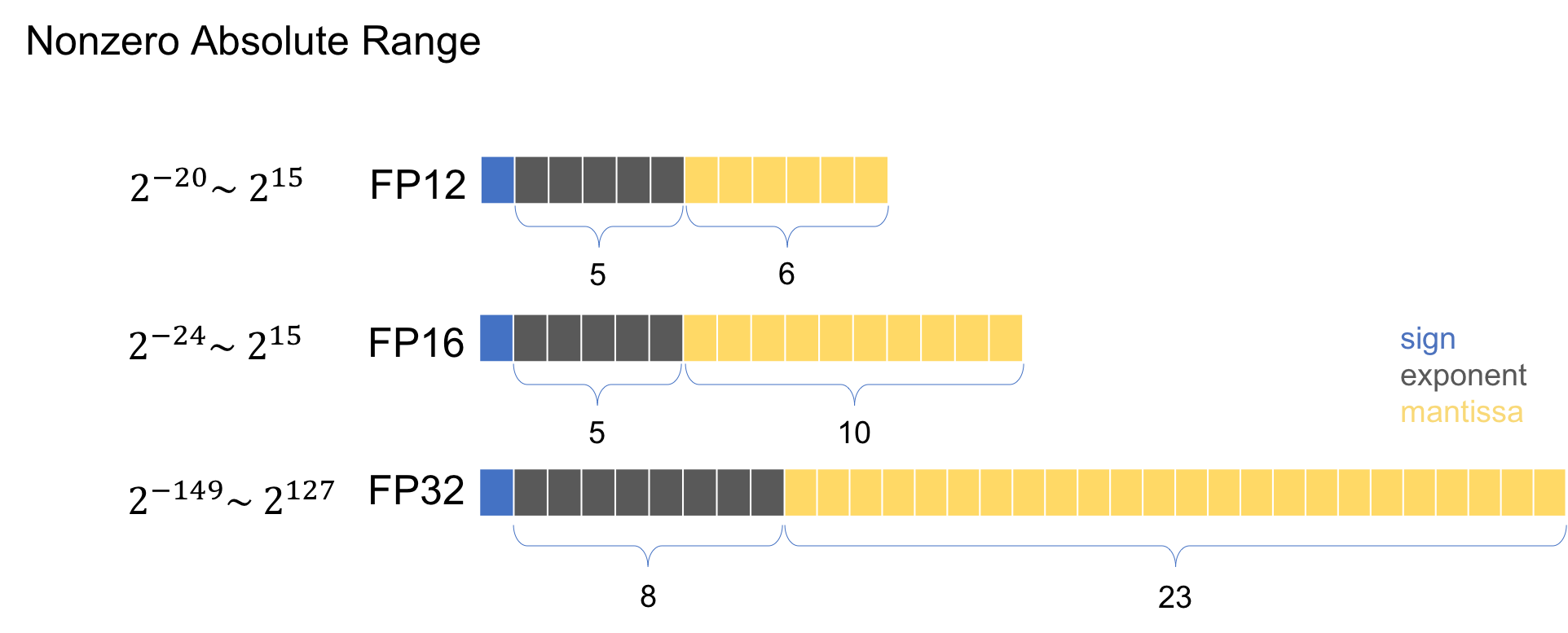}
  \caption{IEEE 754 Floating Point Format\label{fig:ieee_format}}
 \end{figure}
 
 The IEEE Standard for Floating-Point Arithmetic (IEEE 754)~\cite{is2008754} is a technical standard for floating-point computation, which specifies the exchange and arithmetic format and method of binary and decimal floating-point operations in a computer programming environment. The IEEE 754 standard requires support for a handful of operations. These include the arithmetic operations add, subtract, multiply, divide, square root, fused multiply-add, remainder, conversion operations, scaling, sign operations, and comparisons. 
 
 The standard mandates binary floating-point data be encoded on three fields: a one-bit sign field, followed by exponent bits encoding the exponent offset by a numeric bias specific to each format, and bits encoding the significant (or fraction).
 
 To ensure consistent computations across platforms and to exchange floating-point data, IEEE 754 defines basic and interchange formats. However, although Google has introduced a BFloat16 training on its designed chips, Tensor Processing Unit (TPU)~\cite{ying2018image}. However, the design of BFloat16 has sacrificed computational versatility.
 
 \section{Full Low-Precision Training Strategy}
 In this section, we introduce a pure FP16 training strategy which is more memory-efficient than BFloat16 and Float32, without sacrificing training accuracy. We have explored the limitation of low-precision training.
 
 \subsection{Swap the BatchNorm~\label{subsec:swapbn}}
 We have found Mantissa Underflow Risk and Exponent Underflow Risk during the Batch Normalization Layer training process. The underflow exception shall be signaled when a tiny non-zero result is detected.
 
 Batch Normalization(BN)~\cite{ioffe2015batch} is a popular technique that normalizes the activation statistics at the output of every layer, reducing dependencies across layers while significantly improving model accuracy. However, Batch Normalization requires some operations that require high precision and a broad dynamic range, including the computation of the sum of squares, square-root, and reciprocal operations to avoid zero variance, which requires high precision and a broad dynamic range. What is more, we analyze the variance of each layer $x_l $, denoted by $Var[x_l]$ (which is technically defined as the sum of the variance of all the coordinates of $x_l$). For the deep structure like ResNet, prevents $x_l$ from vanishing by forcing the variance to grow with depth, i.e. $Var[x_l ] < Var[x_{l+1}]$ if $E[Var[F(x_l)|x_l ]] > 0$. This causes the output variance to explode exponentially with depth without normalization for positively homogeneous blocks, which are detrimental to learning because it can, in turn, cause gradient explosion.
 
 The most exponent-risky term is the gamma ($\gamma$) in the BN.
 For a layer with $n\times d$ - dimensional input $x=(x_1,x_2,\cdots , x_d)$, the traditional batch normalization would normalizes each dimension
 \begin{equation}
  \hat{x}^{(d)}=\frac{x^{(d)}-\mu^d}{\sqrt{VAR[x^{(d)}]+\epsilon}}
 \end{equation}
 \begin{equation}
  y^{(d)}=\gamma \hat{x}^{(d)}+\beta^{(d)}
 \end{equation}
 
 where the given $\mu^d$ is the expectation of the $x^{(d)}$, $n$ is the batch size, and the $VAR[x^{(d)}]$ is the variance of the given input. $\gamma$ and $\beta$ are a pair of parameters to scale and shift the normalized value. The term $\sqrt{Var[x^{(d)}]}$ involves sums of squares, and the learned that could lead to numerical instability as well as to arithmetic overflow when dealing with large values.
 Since the $\gamma$ is to scale on the related highly unstable term $\frac{1}{\sqrt{Var[x^{(d)}]}}$, the potential of the arithmetic overflow would be enlarged.
 
 To tackle the above challenges, an intuitive solution is to learn the $\frac{1}{\gamma}$ instead of $\gamma$. However, we have found that learning $\frac{1}{\gamma}$ instead of $\gamma$ would improve the training stability with hurting precision, since
 IEEE standard intrinsically defined the product of positive MAX and positive MIN = $2^{bit\;of\;mantissa - 1}$.
 \begin{equation}
  \frac{activation-\beta}{1-\beta}
 \end{equation}
 
 To address this problem, we propose to swap the position of residual connection and gamma ($\gamma$) in BatchNorm after the last convolution layer in each Residual Block means that less processing and less precision requirement when adding up results from $2$ different branches in the computing chip.
 
 \subsection{Mantissa Underflow Risk - Exponent Term Dilemma}
 We found the fact that there is a dilemma between the Mantissa Underflow risk and the Exponent Term that the less Mantissa Underflow risk the more Exponent Term. The Mantissa Underflow is the process of aligning mantissa; digits may flow off the right end of the mantissa. In such a case truncation method such as chopping, the round is used. The mantissa underflow risk comes from the range of weights, which is proven not sensitive in~\cite{zhang2018lq}.
 
 However, we found that it can be easily tackled if introducing LARS-like method.
 
 We do not use LARS, but we multiples output by $0.25$ right after every BatchNorm in each Residual Block. Meanwhile, the BatchNorm in the stem is kept as it is.
 
 \section{Experiments}
 \subsection{Setup}
 In this section, we adopted the ResNet50~\cite{He2016} on ~\cite{deng2009imagenet} a large-scaled GPU cluster with $128$ GPUs. Our framework is PyTorch~\cite{paszke2017pytorch}. Hence, we have adopted an ablation study with $8, 64, 128$ GPU set up with different precision in the same experiment set up. To note that, the learning rate and the batch size has a strong correlation. In our experiment, when the batch size is multiplied by $k$, multiply the learning rate by $k$. All other hyper-parameters (weight decay, etc.) are kept unchanged. Because the training strategy of FP32 has many skills and data enhancement strategies to improve the training effect, to make a fair comparison, the training strategy of Resnet50 in FP32 precision was used from the public repository\footnote{https://github.com/PyTorch/examples/tree/master/imagenet}. All our experiments were conducted under the same settings. For large-scale training, all batch normalization is implemented in a single GPU, which means synchronous batch normalization is not used throughout the pipeline.
 
 The initialization strategy of our experiments is a zero weight decay in all Batch Normalization layers, and zero $\gamma$ in the last Batch Normalization layer in every Residual Block. To avoid a sudden increase in the learning rate, allowing healthy convergence at the start of training, we adopted our learning rate schedule as the same as~\cite{goyal2017accurate}, which we set $5$ round warm-up~\cite{He2016}, decay the learning rate by 0.1 at 30th/60th/80th epochs. The total length of our training is 90 epochs, which are the same as~\cite{goyal2017accurate}.
 
 \subsection{Ablation Study\label{sec:ablation_study}}
 Table~\ref{table:ablation_study} depicts the ablation study of our proposed method. Our baseline (FP32) model is the single-precision storage, which is used for activations, weights, and gradients. All arithmetic is also in FP32. The FP16 precision indicated that all arithmetic and its related activations, weights, and gradients are all stored in FP16 precision. There is no master copy of any parameters that are stored in high precision.
 The experimental results show that the proposed pure FP16 training strategy does not sacrifice much precision. Under the same experimental conditions, compared with the traditional FP32 training strategy, our Top-1\% accuracy is only reduced by 0.7\%, which is acceptable in the large-scale image classification challenge.
 
 Our experiments show it can be harmful to precision when the batch size is over $128\times 32=4096$, and is fatal to convergence when batch size is over $256 \times 32=8192$ if we continuously use the "the batch size is multiplied by $k$, multiply the learning rate by $k$." strategy in~\cite{goyal2017accurate} and this is even cannot be alleviated by using LARS~\cite{you2018imagenet}.
 
 \begin{table}[!ht]
  \centering
  \caption{Large-scale ResNet-50 training results.~\label{table:ablation_study}}
  \begin{tabular}{|c|c|c|c|c|c|c|c|}
  \hline
  Precision & Chips & Batch Size & Top 1(\%) &BN Swap & LARS & Source \\ \hline
  FP32 & 8 & 256 & 75.6 & No & No & Ours \\ \hline
  FP16 & 8 & 256 & 74.7 & No & No & Ours \\ \hline
  FP16 & 64 & 2048 & 75.3 & No & No & Ours \\ \hline
  FP16 & 64 & 2048 & 74.5 & Yes & No & Ours \\ \hline
  FP16 & 128 & 4096 & 74.5 & No & No & Ours \\ \hline
  BFloat16 & 256 & 16384 & 75.08 & No& No & Google~\cite{ying2018image} \\ \hline
  BFloat16 & 256 & 32768 & 76.40 & No& Yes & Google \\ \hline
  % BFloat16 & 1024 & 65536 & 75.20 & No& Yes & Google \\ \hline
  FP32 & 1024&32768& 73.0 & No& Yes &Berkeley~\cite{you2018imagenet} \\ \hline
  \end{tabular}
  \end{table}
 \section{Conclusion}
 In this work, we have analysed the details of the pure FP16 training strategy. We have established that the communication cost reduction, model compression and large-scale distributed training are three coupled problems. This is the first full FP16 training strategy on ResNet50 without sacrificing the accuracy on the ImageNet dataset, which is also the first solution for 5 exponent lower-bound for a wide range of deep learning training schemes.
 \section{Acknowledgement}
 We want to express our special appreciation and thanks to Feng Xiong from Tsinghua University, who has given us advice about the quantization and Yang You from Google Brain, who has given us great support on large-scaled distributed training.

\bibliographystyle{apalike}
\bibliography{refer}

\end{document}